\documentclass[letterpaper, 10 pt, conference]{ieeeconf}

\IEEEoverridecommandlockouts

\usepackage[caption=false, font=footnotesize]{subfig}

\usepackage[acronym]{glossaries}
\usepackage{booktabs}
\usepackage{textcomp}

\usepackage{graphicx}
\usepackage{svg}
\usepackage{subcaption}   
\usepackage{xcolor}
\usepackage{epsfig} 
\usepackage{graphics} 
\usepackage{times} 
\usepackage{amsmath} 
\usepackage{booktabs, multirow}
\usepackage{soul}
\usepackage{changepage,threeparttable} 
\usepackage{amssymb}  
\usepackage{graphicx}
\usepackage{lipsum}  
\usepackage{amssymb}
\usepackage{array}
\usepackage{flushend}
\usepackage[nosort,noadjust]{cite}
\usepackage{amsmath}

\usepackage{float}
\usepackage{microtype}
\usepackage{xargs}
\usepackage{booktabs}
\usepackage{bm}
\usepackage{dsfont}

\usepackage[hidelinks]{hyperref}
\usepackage[capitalise]{cleveref}
\usepackage{float}

\definecolor{myred}{HTML}{ff3030}

\title{\LARGE \bf 
U-Joint CAAMS: Experimental Evaluation of a Universal-Joint Continuum Manipulator for Aerial Manipulation
\vspace{-0.2cm}
}

\author{Anuraj Uthayasooriyan\textsuperscript{1,2}, Musab Alibrahim\textsuperscript{2},  Krishna Manaswi Digumarti\textsuperscript{1,2}, \\ Fernando Vanegas\textsuperscript{1,2},  Felipe Gonzalez\textsuperscript{1,2} 
\thanks{The authors wish to acknowledge the continued support from the Queensland University of Technology (QUT) through the QUT Centre for Robotics, the support of the Research Engineering Facility (REF) team  and the senior technicians  Steven Bulmer, Benjamin Brownlee, and Amir Moghaddam at QUT for the provision of expertise and research infrastructure, and the technical support from Julian Galvez (REF), and P, Keertinathan.}
\thanks{\textsuperscript{1}QUT Centre for Robotics, \textsuperscript{2}School of Electrical Engineering and Robotics, \textsuperscript{3}REF, QUT, Brisbane, QLD Australia 4000.
{\tt\small uthayaso@qut.edu.au | uanuraj@hotmail.com} 
\textsuperscript{$\dagger$}\,Video: \text{\nolinkurl{https://youtu.be/eEp1iFBMUUQ}}}
}

\usepackage{amsmath}
\usepackage{amssymb}
\usepackage{accents}
\usepackage{bm}
\usepackage{xifthen}
\usepackage{color}
\usepackage{fancyvrb}
\usepackage{graphicx,scalerel}

\newcommand{\ba}{\begin{eqnarray}}
\newcommand{\ea}{\end{eqnarray}}

\newcommand{\presup}[1]{\,{}^{\scriptscriptstyle #1}\!}

\newcommand{\pose}[1][ZZZZ]{\ifthenelse{\equal{#1}{ZZZZ}}{}{\presup{#1}}{\mathbf{\xi}}}
\newcommand{\estpose}[1][ZZZZ]{\ifthenelse{\equal{#1}{ZZZZ}}{}{\presup{#1}}{\mathbf{\hat{\xi}}}}
\newcommand{\hpose}[1][ZZZZ]{\ifthenelse{\equal{#1}{ZZZZ}}{}{\presup{#1}}{\hat{\mathbf{\xi}}}}
\newcommand{\posedot}[1][ZZZZ]{\ifthenelse{\equal{#1}{ZZZZ}}{}{\presup{#1}}{\mathbf{\nu}}}

\newcommand{\q}[1][ZZZZ]{\ifthenelse{\equal{#1}{ZZZZ}}{}{\presup{#1}}{\mathring{q}}}

\DeclareMathAlphabet{\mathitbf}{OML}{cmm}{b}{it}
\newcommand{\twist}[2][ZZZZ]{\ifthenelse{\equal{#1}{ZZZZ}}{}{\presup{#1}}{\mathcal{S}}}
\renewcommand{\vec}[2][ZZZZ]{\ifthenelse{\equal{#1}{ZZZZ}}{}{\presup{#1}}{\mathitbf{#2}}}

\newcommand{\hvec}[2][ZZZZ]{\ifthenelse{\equal{#1}{ZZZZ}}{}{\presup{#1}}{\tilde{\vec{#2}}}}
\newcommand{\obvec}[2][ZZZZ]{\ifthenelse{\equal{#1}{ZZZZ}}{}{\presup{#1}}\rlap{${\overbridge{\phantom{$\vec{#2}$}}}$}\vec{#2}}
\newcommand{\evec}[2][ZZZZ]{\ifthenelse{\equal{#1}{ZZZZ}}{}{\presup{#1}}{\hat{\vec{#2}}}}
\newcommand{\bvec}[2][ZZZZ]{\ifthenelse{\equal{#1}{ZZZZ}}{}{\presup{#1}}{\bar{\vec{#2}}}}

\newcommand{\dvec}[2][ZZZZ]{\ifthenelse{\equal{#1}{ZZZZ}}{}{\presup{#1}}{\dot{\vec{#2}}}}
\newcommand{\ddvec}[2][ZZZZ]{\ifthenelse{\equal{#1}{ZZZZ}}{}{\presup{#1}}{\ddot{\vec{#2}}}}

\newcommand{\mat}[2][ZZZZ]{\ifthenelse{\equal{#1}{ZZZZ}}{}{\presup{#1}\,}{{\boldsymbol #2}}}
\newcommand{\dmat}[2][ZZZZ]{\ifthenelse{\equal{#1}{ZZZZ}}{}{\presup{#1}\,}{{\dot{\boldsymbol #2}}}}
\newcommand{\emat}[2][ZZZZ]{\ifthenelse{\equal{#1}{ZZZZ}}{}{\presup{#1}\,}{\hat{\boldsymbol#2}}}
\newcommand{\matfn}[3][ZZZZ]{\ifthenelse{\equal{#1}{ZZZZ}}{}{\presup{#1}}{{\mat{#2}}\left(#3\right)}}
\newcommand{\Rt}[2][ZZZZ]{\ifthenelse{\equal{#1}{ZZZZ}}{}{\presup{#1}}{{\bf R}\left(#2\right)}}

\newcommand{\point}[2][ZZZZ]{\ifthenelse{\equal{#1}{ZZZZ}}{}{\presup{#1}}{\mathbf{\mathrm{#2}}}}

\newfont{\School}{pncr}
\newfont{\eightTR}{pncr at 8pt}

\usepackage{color}
\usepackage{fancyvrb}
\fvset{formatcom=\color{blue},fontseries=c,fontfamily=courier,xleftmargin=4mm,commentchar=!}

\DefineVerbatimEnvironment{Code}{Verbatim}{formatcom=\color{blue},fontseries=c,fontfamily=courier,fontsize=\footnotesize,xleftmargin=4mm,commentchar=!}

\DefineVerbatimEnvironment{CodeSmall}{Verbatim}{formatcom=\color{blue},fontseries=c,fontfamily=courier,fontsize=\scriptsize,xleftmargin=1mm,commentchar=!}

\DefineVerbatimEnvironment{CodeNum}{Verbatim}{numbers=left,numbersep=4pt,formatcom=\color{blue},fontseries=c,fontfamily=courier,fontsize=\footnotesize,xleftmargin=4mm}

\newcommand{\model}[1]{\index{code}{#1@\textit{#1}}\ifthenelse{\boolean{draft}}{{\color{green}\Verb+#1+}}{\Verb+#1+}}
\newcommand{\block}[1]{\ifthenelse{\boolean{draft}}{{\color{green}\Verb+#1+}}{\textsf{#1}}}

\newcommand{\func}[2][ZZZZ]{\ifthenelse{\equal{#1}{ZZZZ}}{\index{code}{#2}}{\index{code}{#1}}\ifthenelse{\boolean{draft}}{{\color{green}\Verb+#2+}}{\Verb+#2+}}

\newcommand{\methodb}[2]{\index{code}{#1@\textbf{#1}!.#2}\ifthenelse{\boolean{draft}}{{\color{magenta}\Verb+#1.#2+}}{\Verb+#1.#2+}}
\newcommand{\method}[2]{\index{code}{#1@\textbf{#1}!.#2}\ifthenelse{\boolean{draft}}{{\color{magenta}\Verb+#2+}}{\Verb+#2+}}
\newcommand{\class}[1]{\index{code}{#1@\textbf{#1}}\ifthenelse{\boolean{draft}}{{\color{cyan}\Verb+#1+}}{\Verb+#1+}}
\newcommand{\property}[1]{\index{property}{#1}\ifthenelse{\boolean{draft}}{{\color{cyan}\Verb+#1+}}{\Verb+#1+}}

\begin{document}

\makeatletter
\let\@oldmaketitle\@maketitle
\renewcommand{\@maketitle}{\@oldmaketitle
\centering
\setlength{\belowcaptionskip}{-8pt}
\includegraphics[width=17.5cm]{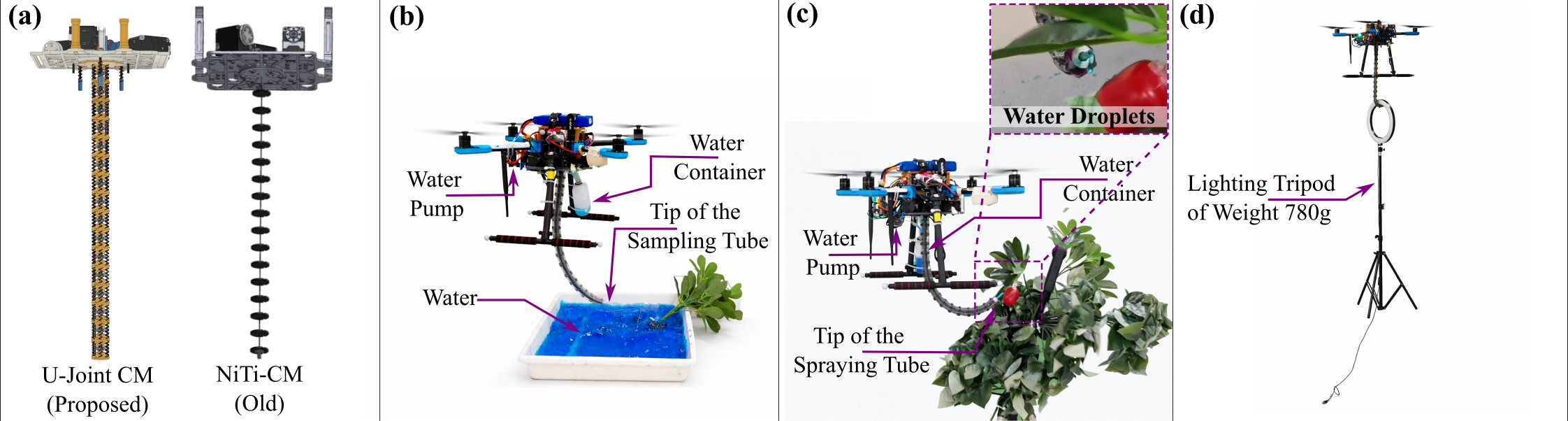}
\vspace{-0.1cm}
\captionof{figure}{(a) CAD versions of the proposed U-Joint CM  and  prior NiTi CM, (b)in flight water sampling , (c)spot spraying on an artificial plant target, and (d) whole-body grasping and transportation performed by the U-Joint CAAMS (See the accompanying video\textsuperscript{$\dagger$}). }
\label{fig:main_figure}
\vspace*{-6pt}
}
\makeatother

\maketitle
\setcounter{figure}{1}
\renewcommand{\thefigure}{\arabic{figure}}

\begin{abstract}

Continuum manipulators mounted on multi-rotor UAVs enable compliant aerial manipulation, but payloads and propeller downwash amplify out-of-plane bending and twisting that degrade end-effector pose accuracy. To address this problem, we present a universal-joint-based continuum manipulator designed to improve resistance to out-of-plane deformation during aerial manipulation. The proposed design uses a tubular backbone with spring-reinforced universal joints and an integrated conduit for internal routing and fluid delivery. We evaluate the design in still air and under peak propeller downwash across varying payloads, and benchmark it against a prior Nitinol-backbone CM. Bench tests show improved resistance to out-of-plane deformation across all conditions. Under peak downwash, the proposed design reduces mean error by $2.5$--$4\times$ in yaw, $2$--$45\times$ in $y$-axis, and up to $5\times$ in roll compared to the NiTi-backbone design. We further analyze hover stability through in-flight coupled-disturbance tests over varying payloads and actuation speeds, and demonstrate the system in water sampling, spot spraying, and object transport.

\end{abstract}


\section{Introduction}

Multi-rotor uncrewed aerial vehicles (UAVs) equipped with robotic arms can interact with objects in flight, enabling tasks ranging from sensing to physical manipulation. The applications include infrastructure monitoring\cite{Hamaza2020}, environmental sample recovery\cite{sanim2022development},\cite{geckeler2025field} sensor deployment and retrieval\cite{Hamaza2020}, and spot spraying\cite{delavarpour2023review}. 
Continuum manipulators (CM) are increasingly preferred over rigid link robot arms for their light weight, better dexterity and compliance \cite{szasz2022modeling, peng2023aecom, peng2025dexterous, jalali2022aerial, uthayasooriyan2026tiltxenablingcompliantaerial}. Manipulator inertia influences coupled disturbance (CPD) of aerial manipulators \cite{xu2025atom}. Continuum arm aerial manipulator systems (CAAMS) also have the advantage of low manipulator inertia as the actuators can be remotely located within the UAV \cite{jalali2022aerial}.  

Due to softness, CMs have difficulty in tracking a desired end effector (EE) pose. In applications of spraying, liquid sampling and sensing, the CM must carry extra load. Further, the downwash (DW) from the UAV-propellers exerts aerodynamic loads on the CM. This combination of external loads worsens the pose-tracking of CMs. Out of plane bending, amplified by external loads has been found as a major contributor to the pose error as observed in our prior work \cite{uthayasooriyan2026tiltxenablingcompliantaerial}.

 \begin{figure*}[htb]
    \vspace{4pt}
    \centering
    \includegraphics[width=\linewidth]{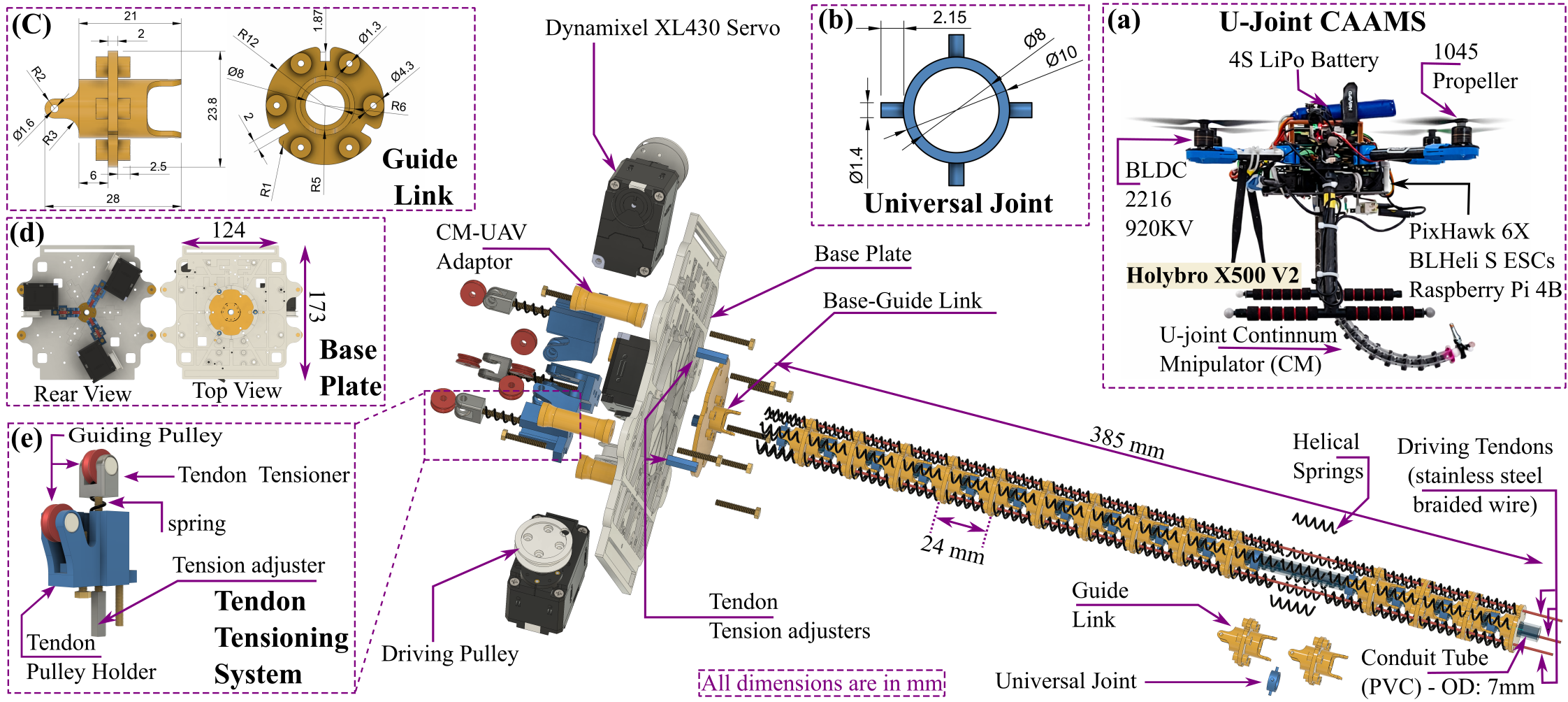}
    \caption{The design details of the proposed U-Joint CM: (a) The U-Joint CAAMS prototype, (b) Designed universal joint , (c) Tubular guide link, (d) The actuation compartment base plate, and (e) Tendon tensioning system.} 
    \label{fig:U-joint_Exploded}
    \vspace{-12pt}
\end{figure*}

CAAMS research has emerged only in recent years. More studies focus on dynamics and control, and comparatively less on design \cite{peng2023aecom, peng2025dexterous, chien2021kinematic, chien2023design, jitosho2025flying}. Samadikhoshkho et al. did foundational work on CAAMS, including coupled modeling and vision-based control \cite{samadikhoshkhocoupled, samadikhoshkho2020modeling, samadikhoshkho2022vision}. A dynamics controller for a dual-arm CAAMS was presented in \cite{ghorbani2023dual}, and integration/control of a soft CM on an omnidirectional micro multi-rotor was explored in \cite{szasz2022modeling}. Beyond these, recent efforts have reported prototyping and control contributions, including mechanical analysis of modular CMs \cite{zhao2022modular}, tethered CAAMS designs \cite{chien2021kinematic,chien2023design}, tendon slack prevention with IMU-based closed-loop control\cite{peng2023aecom}, a  3--section CM--CAAMS for cluttered environments\cite{peng2025dexterous}, vine CM \cite{jitosho2025flying} and tiltable and extensible CM system targeted for deployment while avoiding DW \cite{uthayasooriyan2026tiltxenablingcompliantaerial}.

Although, \cite{uthayasooriyan2026tiltxenablingcompliantaerial} reports DW-amplified orientation errors, a design-oriented solution that robustly mitigates out-of-plane deformation remains largely unexplored. From an aerial manipulation perspective, Cartesian position errors can often be compensated by leveraging the UAV motion. In contrast, orientation errors induced by out-of-plane bending or twisting of the CM are comparatively harder to correct.

Inspired by \cite{peng2023aecom, GrassmannBurgner-Kahrs_et_al_Frontiers_2023, clark2020continuum, yeshmukhametov2022development, uthayasooriyan2024tendon}, in this work, we present a tubular, universal-joint CM (U-joint CM) design with a flexible conduit backbone (Fig.1(a) and Fig.2). The primary focus is to improve the out of plane bending errors observed in the Nitinol  backbone CM (NiTi - CM) design similar to the one used in Tilt-X\cite{uthayasooriyan2026tiltxenablingcompliantaerial}. As discussed in our prior review \cite{uthayasooriyan2024tendon}, we chose a universal-joint profile to improve torsional resistance without the added weight and actuation complexity of stiffness-tuning approaches such as phase-change materials \cite{wang2022design}, curvature-constraining rods \cite{zhao2018continuum}, and jamming-based methods \cite{quteprints264715,howard2023comprehensive,uthayasooriyan2024tendon}. Springs reinforce the universal joints to improve rigidity and elasticity against out-of-plane bending and twisting. We experimentally evaluate the proposed design in (1)still air and (2)under propeller DW across varying payloads, and benchmark its behavior against the continuum design in \cite{uthayasooriyan2026tiltxenablingcompliantaerial}. Results show improved resistance to out-of-plane bending. We further assess U-joint CAAMS's hover stability via in-flight coupled-disturbance (CPD) analysis during manipulator actuation across multiple speeds and payloads. We also demonstrate the proposed U-joint CAAMS in three in-flight tasks: water sampling, spot spraying, and object transport as shown in Fig 1(b) -- Fig 1(d).

In summary, this work presents the:
\begin{enumerate}
        \item Design of a tubular universal-joint-based CM with an integrated conduit backbone for multi purpose aerial manipulation.
        \item Experimental evaluation of the design for out of plane bending and twisting under varying payloads in still air and propeller-downwash.
        \item Quantitative analysis of in-flight hover stability via coupled-disturbance (CPD) during CM actuation.
        \item  Demonstration of the system in three in-flight tasks.
        
\end{enumerate}

\section{Design and System Architecture}
\label{sec:System_Architecture}

\subsection{Design of U-Joint continuum manipulator (U-Joint CM)} 

This design incrementally improves the continuum section used in our prior Tilt-X work. Motivated by Tilt-X limitations and the target aerial manipulation scenarios in Sec.~I, we define the following U-joint CM design requirements.

\begin{enumerate}
    \item Reducing out of plane bending.
    \item Increasing payload handling capacity.
    \item Increasing stability against the propeller DW.
    \item Adaptable for the applications of environment water sampling, spraying and carrying sensors.
\end{enumerate}


In Tilt-X, out-of-plane bending was primarily driven by torsion of the NiTi backbone (backbone diameter: 1\,mm, tendons: braided micro filament threads). Here we replace it with a tubular universal-joint backbone reinforced by radially arranged chain of helical springs to increase torsional stiffness and provide enough elasticity for shape restoration. \Cref{fig:U-joint_Exploded} shows the CAD exploded view and key dimensions. The tubular joint--link geometry integrates a concentric PVC conduit, selected for its flexibility and ability to recover the straight reference shape. The tendons are 1\,mm braided steel cables. Key dimensions (disk spacing, tendon routing hole(eyelet) distance, length, and outer diameter) match the Tilt-X NiTi CM to ensure a fair experimental comparison. Except for the springs, tendons, routing pulleys, and fixtures, all parts are 3D-printed in PLA. The actuation compartment uses three daisy-chained Dynamixel XL430 servos driven via a U2D2 interface and Power Hub. Spring-loaded pretensioners with tension adjusters set the initial tendon tension and shape of the U-joint CM. The U-joint CM weighs 430\,g, whereas the NiTi CM weighs 330\,g including actuation compartment. The total mass of the U-joint CAAMS, including the pump and battery, is 2350\,g.

{%
\begin{figure}[htb]
    \vspace{6pt}
    \centering
    \includegraphics[width = \linewidth]{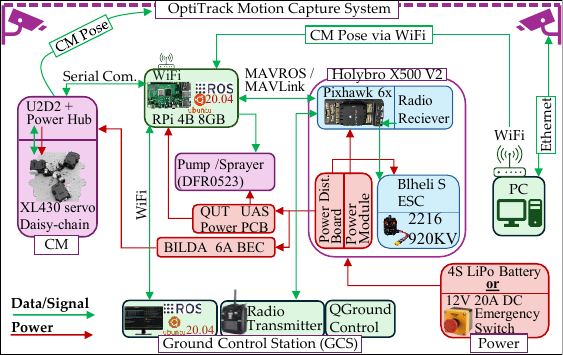}
    \caption{The system architecture of the U-Joint CAAMS proposed and experimented under bench test and flight tests.}
    \label{fig:System_archi}
    \vspace{-1\baselineskip}
\end{figure}
}%

\subsection{System architecture of proposed U-Joint CAAMS} 


\cref{fig:System_archi} summarizes the U-joint CAAMS architecture. The platform comprises a Holybro X500~V2 quadrotor with the proposed U-joint CM (see \cref{fig:System_archi}; quadrotor specs in \cref{fig:U-joint_Exploded}(a)). Power is supplied by a 4S LiPo battery for flight and a 12\,V/20\,A DC supply for bench tests. A Raspberry Pi~4B runs ROS to control the CM and interfaces with the flight controller via MAVROS/MAVLink. The onboard computer is accessed from a ROS-enabled ground station, while QGroundControl and an RC transmitter provide flight monitoring and command. System pose is measured using OptiTrack motion capture system. For water sampling and spot spraying, a 6\,V/1.8\,A peristaltic pump (DFRobot DFR0523, 85\,ml/min) is controlled by the Raspberry Pi.


\subsection{Kinematic model} 
\label{subsec:Kinematic model}
To enable a fair comparison with the Tilt-X NiTi CM, we adopt the same constant-curvature (CC) model used in \cite{uthayasooriyan2026tiltxenablingcompliantaerial}, following the standard formulation in \cite{webster2010design}. A brief summary is provided below.

\begin{figure*}[!ht]
    \vspace{4pt}
    \centering
    \includegraphics[width=\linewidth]{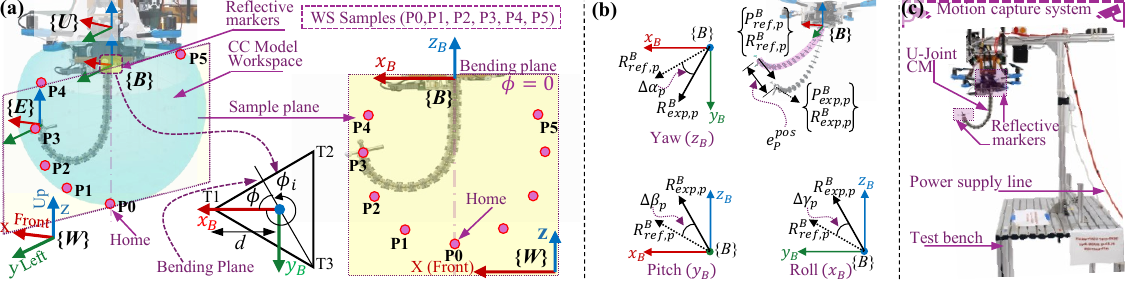}
    \caption{(a)The constant–curvature CM workspace of the CAAMS and reference frames: world $\{W\}$, UAV $\{U\}$, CM base $\{B\}$, and CM tip$\{E\}$. The  triangle indicates tendon locations (T1–T3). The pink dots on the sample plane indicate the sample points P0(Home),P1, P2, P3, P4 and P5 for the experiment.
    (b) Definition of position and orientation errors between the expected $\{{ref}\}$ and the measured $\{{exp}\}$ pose expressed in $\{B\}$. (C) Experiment rig used for the bench-based experiments.
    } 
    \label{fig:workspace}
    \vspace{-10pt}
\end{figure*}

The \cref{fig:workspace}(a) depicts the U-joint CAAMS workspace with the right handed coordinate frame definitions as follow: $\{W\}$: world, $\{U\}$: UAV, $\{B\}$: base of the U-joint CM, and $\{E\}$: tip of the U-joint CM. The frame $\{B\}$ is defined such that the U-joint CM at its rest state aligns with -z axis, tendon $T1$ aligns with the x-axis of the frame $\{B\}$. Further, the x-axis of $\{B\}$ and $\{U\}$ are aligned with the forward direction of the UAV. In the rest state of the  U-joint CM, the frame $\{E\}$ is also designated similar to that of $\{B\}$. End effector pose can be obtained by the following homogeneous transformation:
\vspace{-0.1cm}
{%
\begin{equation}
\label{eq:WorldToEE}
    \mathbf{T}_{WE} = \mathbf{T}_{WU} \cdot \mathbf{T}_{UB} \cdot \mathbf{T}_{BE} .
\end{equation}
}%
\vspace{-0.5cm}

For bench experiments ${T}_{WB}$ is determined directly using the Optitrack motion capture system. When flight test is performed ${T}_{WU}$ is found using the motion capture system while ${T}_{UB}$ is estimated from the design.

Under the assumption of inextensible backbone, a CC model for a CM can be established using the parameters curvature: $\kappa$, bending plane angle: $\phi$ measured from positive x--axis and length of the backbone: $\ell$. Thus the pose of the end effector ${T}_{BE}$ in frame $\{B\}$ is

{%
\setlength{\abovedisplayskip}{-6pt}%
\begin{equation}
\label{eq:CM_homogeneous}
T_{BE} =
\begin{bmatrix}
c_{\kappa\ell} c_\phi & s_\phi & s_{\kappa\ell} c_\phi & \frac{1}{\kappa}(1-c_{\kappa\ell})c_\phi \\
c_{\kappa\ell} s_\phi & -c_\phi & s_{\kappa\ell} s_\phi & \frac{1}{\kappa}(1-c_{\kappa\ell})s_\phi \\
s_{\kappa\ell} & 0 & -c_{\kappa\ell} & -\frac{1}{\kappa}s_{\kappa\ell} \\
0 & 0 & 0 & 1
\end{bmatrix}.
\end{equation}
}%
where $c_\phi = \cos\phi$, $s_\phi = \sin\phi$, $c_{\kappa\ell} = \cos({\kappa\ell})$, and $s_{\kappa\ell} = \sin({\kappa\ell})$.

As the design imply, the tendons are distributed at equal spacing around the backbone of length $L$ at a radial distance $d$. For this continuum section, the kinematic parameters can be presented in terms of the tendon--length actuation vector $q = [l_{1}, l_{2}, l_{3}]^T$ by the following equations:

{%
\vspace{-0.6\baselineskip}   
\begin{equation}
\label{l_q}
\ell(q) = L
\end{equation}
}%

{%
\vspace{-0.9\baselineskip}   
\begin{equation}
\label{BendingPlaneAngle}
\phi(q) = \operatorname{atan2}\!\left(\sqrt{3}\,(l_{2}-l_{3}),\, l_{2}+l_{3}-2l_{1}\right)
\end{equation}

}%

{%
\vspace{-0.9\baselineskip}   
\begin{equation}
\label{Kappa}
\kappa(q) = \frac{2}{d\,(l_{1} + l_{2} + l_{3})}
\sqrt{\,l_{1}^{2} + l_{2}^{2} + l_{3}^{2} - l_{1}l_{2} - l_{1}l_{3} - l_{2}l_{3}}
\end{equation}
\vspace{-0.6\baselineskip}
}%

where $l_{i}$ is the length of each tendon with $i=1,2,3$. 
To actuate the CM, the required change  in tendon length $\Delta l_{i}$ of a tendon $l_{i}$ to result a change in curvature and bending in a targeted plane is

{%
\vspace{-0.8\baselineskip}   
\begin{equation}
\label{continuum_tendon_Del_L}
\begin{aligned}
\Delta l_{i} &= -L \kappa d \cos\left( \phi_{i}\right), \\
\end{aligned}
\end{equation}
}%

where $\phi_{i}$ is the location of the tendon relative to the bending plane at the base of the CM. The effective change of $\Delta l_{i}$ is created by the rotation of the driving pulleys of radius ${r}_{\text{m}_i}$ coupled to the servo motors in the actuation compartment, applying a turn of ${\psi}_{i}$ can then be written as

{%
\vspace{-0.6\baselineskip} 
\begin{equation}
    \label{Del_L_motor}
    {\Delta l_{i} = {r}_{\text{m}_i}{\psi}_{i}},
\end{equation}
}%

\section{Experiments} 
\label{sec:Bench_Experiments}

\subsection{Experiment Design}

We conducted three experiments: (i) bench evaluation of U-joint CM, (ii) flight hover testing for CPD following \cite{xu2025atom}, and (iii) in-flight application demonstrations. \Cref{tab:experimentDesign} summarizes the experimental plan. Bench tests were conducted under three cases: (i) no DW (props off - T0) for U-joint vs NiTi comparison, (ii) no DW with variable end-effector payload, and (iii) maximum-throttle (T100) assuming a worst case scenario with variable payload. Payloads were swept from 0 to 50\,g in 10\,g increments. We assume that maximum EE loading of 50\,g is reasonable considering a light weight sensor of the amount of liquid that can stay within the U-Joint conduit.

    
{%
\begin{table}[h]
    \centering
    \caption{Bench-top experiments under different cases}
    \label{tab:experimentDesign}
    \setlength{\tabcolsep}{2pt}  
    \resizebox{\columnwidth}{!}{%
    \begin{tabular}{>{\raggedright\arraybackslash}p{0.15\linewidth}>{\raggedright\arraybackslash}p{0.25\linewidth}>{\raggedright\arraybackslash}p{0.25\linewidth}>{\raggedright\arraybackslash}p{0.40\linewidth}}
        \toprule
        Cases & Throttle T\% & EE Loading (g) & Physical interpretation \\
        \midrule
        Case(i)   & 0   & 0                    & No DW (NiTi Vs U-Joint) \\
        Case(ii)  & 0   & 0,10,20,30,40,50    & No DW; Variable EE loading \\
        Case(iii) & 100 & 0,10,20,30,40,50    & DW; Variable EE loading \\
        \bottomrule
    \end{tabular}}
\end{table}
\vspace{-0.9\baselineskip}
}%

\subsection{Experiment Setup}
The \cref{fig:workspace}(c) depicts the experiment bench setup. For bench experiments, each type of CAAMS was rigidly mounted to a test bench to constrain all six UAV DOF. An OptiTrack\textsuperscript{\textregistered} motion-capture system tracked reflective markers to define the frames $\{U\}$, $\{B\}$, and $\{E\}$. The CPD hover experiment was performed with the U-joint CAAMS in autonomous position-hold mode.


\subsection{Data Collection}

\subsubsection{Bench experiment}

We first sample the CC-model workspace generated by \cref{eq:CM_homogeneous} (bending $\le 180^\circ$ to avoid propeller interference). \cref{fig:workspace}(a) shows the forward-aligned bending plane with the forward direction of the CAAMS selected for evaluation. We chose four random WS sample points that cover the bending angles from 0--180$^\circ$ to compare NiTi and U-joint CMs. Using the kinematic model (\cref{subsec:Kinematic model}), we precompute tendon length commands for each pose. For each case in \cref{tab:experimentDesign}, the CAAMS is actuated with these commands and all frame poses and actuation variables are logged to ROS \texttt{.bag} files. Payloads are applied in 10\,g steps. Each WS point is measured over three cycles with a 10\,s hold. Preliminary trials showed that the NiTi CM could not reliably hold the commanded poses under a 10g payload; therefore, only the U-joint CM was tested for the payload sweep.

\subsubsection{Flight hover experiment}
Following \cite{xu2025atom}, we evaluate hover stability of U-Joint CAAMS via coupled-disturbance (CPD) analysis during sinusoidal U-joint CM motion in autonomous position hold at $(0,0,1.5)$\,m. We record U-Joint CAAMS's position in two conditions: (i) hover only (no manipulator motion) for benchmarking a baseline and (ii) hover with oscillatory CM actuation. Tests are repeated for three payloads: W0, W30, and W50. Each oscillation cycle starts from the home position $P0$ (\cref{fig:workspace}(a)) and executes: home$\rightarrow$P4$\rightarrow$home$\rightarrow$P5$\rightarrow$home. We test six cycle times, $T\in\{2,4,8,12,16,20\}$\,s. The quarter-cycle time ($T/4$) is the travel time from home to an extreme point, consistent with \cite{xu2025atom}. The quarter-cycle time of last three periods match \cite{xu2025atom}; the rest of the shorter periods probe more aggressive, worst-case scenarios for U-Joint CAAMS.


\subsection{Evaluation of data}

To support the analysis in \cref{sec:Results}, we summarize the evaluation metrics in this section. For each WS sample point $P$, we express the pose errors in the CM base frame $\{B\}$ as the Euclidean position error $e_{p}^{\mathrm{pos}}$ given in \eqref{eq:pos_error} and the scalar orientation error $e_{R,p}^{\circ}$ expressed by \eqref{eq:ori_error}. We define $e_{p}^{\mathrm{pos}}$ (mm) as the Euclidean distance between the measured and reference positions and $e_{R,p}^{\circ}$ (deg) as the principal rotation angle of the relative rotation $\mathbf{R}_{\mathrm{rel},p}$, i.e., the axis--angle rotation about unit axis $\hat{\mathbf{u}}_{p}$ that maps the measured orientation to the reference orientation.

{%
\begin{figure*}[htb]
    \vspace{6pt}
    \centering
    \includegraphics[width = \linewidth]{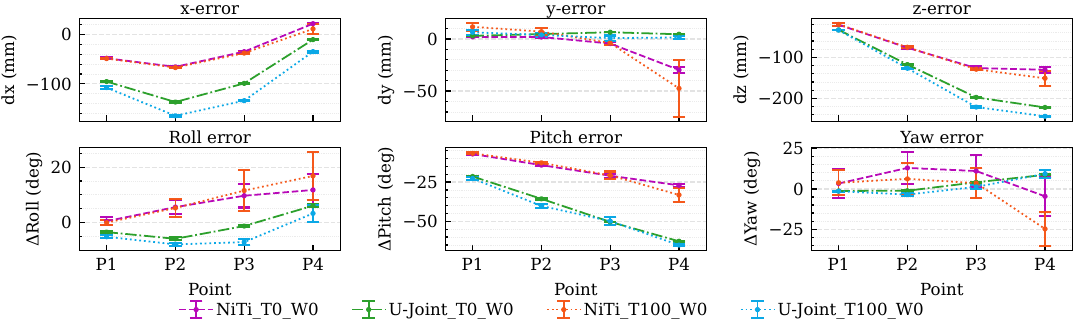}
    \caption{The component wise error analysis plots of U-Joint CM and the NiTi-CM against the Constant Curvature model under no DW effect (Throttle T=0) and no end effector loading. Upper raw shows positional error components and the lower raw depicts orientation error components with rescpect to the CM base. }
    \label{fig:T0W0_ModelError}
    \vspace{-0.9\baselineskip}
\end{figure*}
}%

To analyze Case~(i) in Section IV, the reference pose is provided by the CC-model at the corresponding WS-point. For the rest of the cases, the reference at each WS-point is the mean over repeated measurements from case~(i) : positions are averaged linearly and orientations are averaged as unit quaternions using Markley’s method~\cite{markley2007quaternion}. Euler angles are not averaged; Euler error components are obtained from \eqref{eq:euler_rel_zyx}. For Case~(i) (Sec.~IV), the reference pose at each WS point is the CC-model prediction. For the remaining cases, the reference at each WS point is the Case~(i) mean over repetitions: positions are averaged arithmetically, while orientations are averaged in quaternion space using Markley’s method~\cite{markley2007quaternion}. Euler error components are calculated using \eqref{eq:euler_rel_zyx}.

\subsubsection{Position error}
As seen in \cref{fig:workspace}(a),(b), at each WS sample $P$,  let $\mathbf{P}^{B}_{\mathrm{exp}}=[x^{B}_{\mathrm{exp},p},\,y^{B}_{\mathrm{exp},p},\,z^{B}_{\mathrm{exp},p}]^{\top}$ be the measured EE position for the sample, and $\mathbf{p}^{B}_{\mathrm{ref}}=[x^{B}_{\mathrm{ref},p},\,y^{B}_{\mathrm{ref},p},\,z^{B}_{\mathrm{ref},p}]^{\top}$ be the respective reference EE position. Now, the Euclidean position error can be written as

{%
\begin{equation}
e_{p}^{\mathrm{pos}}
=
\sqrt{
\Delta x_{p}^{2}+\Delta y_{p}^{2}+\Delta z_{p}^{2}
},
\label{eq:pos_error}
\end{equation}
\vspace{0.1\baselineskip}
}%

where $\Delta x_{p}=x^{B}_{\mathrm{exp},p}-x^{B}_{\mathrm{ref},p}$, $\Delta y_{p}=y^{B}_{\mathrm{exp},p}-y^{B}_{\mathrm{ref},p}$, and $\Delta z_{p}=z^{B}_{\mathrm{exp},p}-z^{B}_{\mathrm{ref},p}$.

\subsubsection{Orientation error}

Let $\mathbf{R}^{B}_{\mathrm{exp},p}\in SO(3)$ and $\mathbf{R}^{B}_{\mathrm{ref},p}\in SO(3)$ be the measured and reference EE orientations at the same workspace point. Following, \cite{lynch2017modern}, defining the relative rotation between them:

Let $\mathbf{R}^{B}_{\mathrm{exp},p}\in SO(3)$ and $\mathbf{R}^{B}_{\mathrm{ref},p}\in SO(3)$ denote the measured and reference end-effector orientations at the same WS point. Refering \cite{lynch2017modern}, we define the relative rotation as:

\begin{equation}
\mathbf{R}_{\mathrm{rel},p}
=
\left(\mathbf{R}^{B}_{\mathrm{exp},p}\right)^{-1}\mathbf{R}^{B}_{\mathrm{ref},p}.
\label{eq:rel_error}
\end{equation}

The scalar orientation error is the principal rotation angle associated with $\mathbf{R}_{\mathrm{rel},p}$:

\begin{equation}
e_{R,p}^{ori} = \angle\!\left(\mathbf{R}_{\mathrm{rel},p}\right)\in[0,\pi],
\qquad
e^{\circ}_{R,p}=\frac{180}{\pi}\,e_{R,p}^{ori},
\label{eq:ori_error}
\end{equation}

where $\angle(\cdot)$ returns the principal rotation angle and $e^{\circ}_{R,p}$ is expressed in degrees. For qualitative interpretation, we further decompose $\mathbf{R}_{\mathrm{rel},p}$ into extrinsic (fixed-axis) Z--Y--X Euler angles \cite{scipy_as_euler, scipy_from_euler}:

{%
\vspace{-0.9\baselineskip}
\begin{equation}
\begin{bmatrix}
\Delta\alpha_{p} & \Delta\beta_{p} & \Delta\gamma_{p}
\end{bmatrix}^{\!\top}
=
\mathrm{Euler}_{\text{extrinsic }zyx}\!\left(\mathbf{R}_{\mathrm{rel},p}\right),
\label{eq:euler_rel_zyx}
\end{equation}
}%

where $\Delta\alpha$, $\Delta\beta$, and $\Delta\gamma$ denote yaw--pitch--roll corrections about the fixed base-frame axes $z_B$, $y_B$, and $x_B$, respectively (see \cref{fig:workspace}(b)). Each angle is wrapped to $[-180^{\circ},180^{\circ}]$ to avoid discontinuities. Here, yaw is rotation about $z_B$, pitch is the tilt about $y_B$, and roll about $x_B$.

\subsubsection{Mean and standard deviation (SD) used in plots}

For each workspace point $P$ and the samples $k$, the plotted mean and SD are computed from the set of available scalar errors $\{e_{p,k}\}_{k=1}^{N_{p}}$ (either $e_{p}^{\mathrm{pos}}$ in \eqref{eq:pos_error} or $e^{\circ}_{R,p}$ in \eqref{eq:ori_error}) as
\vspace{-0.4cm}

\begin{align}
\mu_{p} &= \frac{1}{N_{p}}\sum_{k=1}^{N_{p}} e_{p,k}, \label{eq:mean}\\
\sigma_{p} &= \sqrt{\frac{1}{N_{p}}\sum_{k=1}^{N_{p}}\left(e_{p,k}-\mu_{p}\right)^{2}},
\label{eq:std}
\end{align}

where $N_{p}$: number of samples pooled for point $p$.

\section{Results and Discussion}
\label{sec:Results}

As described in \cref{sec:Bench_Experiments}, here we report results from bench experiments, CPD tests, and application demonstrations. Bench tests assess U-joint CM's resistance to out-of-plane bending in still air (T0) and propeller DW (T100) using three cases: (i) benchmarking of both CMs (U-joint vs NiTi) against CC-model , (ii) payload-induced pose variation of the U-joint CM in still air, and (iii) DW-induced pose error at T100 across payloads. Both the Cases (ii)$\&$(iii) use baseline from Case~(i).  We further analyze robustness to out-of-plane bending; the goal is not evaluating CC-model accuracy, but resistance of U-Joint CM to out-of-plane deformation.

Results are analyzed via component-wise errors in position $d_x, d_y, d_z$ and orientation $\Delta$Yaw, $\Delta$Pitch, and $\Delta$Roll. Legends denote \textbf{W} as the EE payload mass (e.g., \textbf{W10} = 10 g) and \textbf{T} as the throttle level (\textbf{T0}: still air; \textbf{T100}: max. throttle). Out-of-plane deformation is captured by paired signatures of($\Delta$Yaw, $d_y$) and ($\Delta$Roll): larger ($\Delta$Yaw, $d_y$) indicates greater out-of-plane bending, while larger ($\Delta$Roll) indicates increased twisting. In-plane bending is reflected by $(\Delta\theta, d_x)$ and $d_z$.


\subsection{Case (i): Benchmarking U-joint and NiTi CMs against the CC--model (still air $T=0$)}
\label{sec:proto vs CC}

\cref{fig:T0W0_ModelError} reports model-relative errors at the workspace (WS) points (P1–P4) for the two prototypes under no payload (W0) with or without DW. Out-of-plane bending is assessed via the paired signatures $(\Delta\text{Yaw},\,d_y)$, while twisting is assessed via $\Delta$Roll. In still air(T0) , the U-joint CM keeps $\Delta\text{Yaw}$ tightly bounded (-4$^\circ$ and +9$^\circ$) with small SD and $d_y$ remains close to zero across points. The NiTi CM shows larger $\Delta\text{Yaw}$  SD and a clear drift in $d_y$, most pronounced at P4 with approximately 4-times mean error at T100 than U-Joint CM. Under DW \text{T100\_W0}, NiTi errors increase sharply (notably $\Delta$Yaw, $d_y$ at P4), whereas the U-joint CM response remains bounded. For the U-joint CM, $\Delta$Roll is similar in T0 and T100, with mean $|\Delta$Roll$|$ bounded within $\approx$12$^\circ$ and a smaller SD. The NiTi CM shows larger twisting, with mean $|\Delta$Roll$|$ $\approx$22$^\circ$ which is around 2x of error and more than double the SD. 


{%
\begin{figure*}[!htb]
    \vspace{6pt}
    \centering
    \setlength{\belowcaptionskip}{-8pt}
    \includegraphics[width = \linewidth]{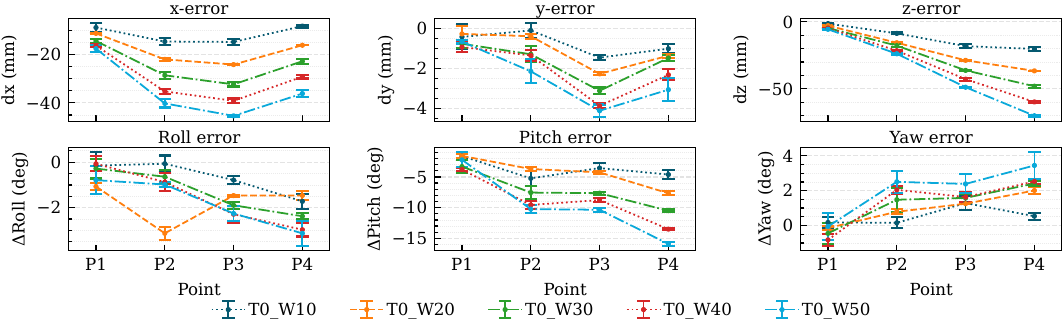}
    \caption{The component wise error analysis plots of U-Joint CM  against the baseline under no DW effect (Throttle T=0) and varying end effector loading of 0g to 50g (E.g. in T0\_W10, T0: no throttle, and W10: 10g load). Upper raw shows positional error components and the lower raw depicts orientation error components with rescpect to the CM base.}
    \label{fig:T0_Load}
\end{figure*}
}%

{%
\begin{figure*}[htb]
    \vspace{6pt}
    \centering
    \includegraphics[width = \linewidth]{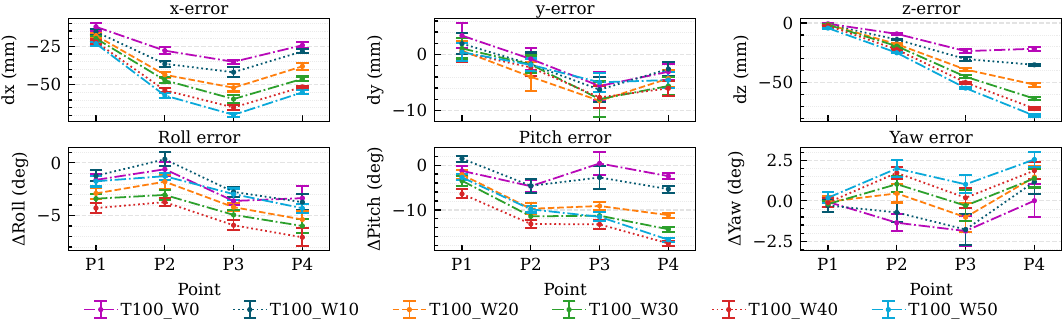}
    \caption{The component wise error analysis of U-Joint CM  against the baseline under extreme DW effect (Throttle T=100\%) and varying end effector loading of 0g to 50g (E.g. in T100\_W10, T100: 100\% throttle, and W10: 10g load). Upper raw shows positional error components and the lower raw depicts orientation error components with rescpect to the CM base.}
    \label{fig:T0_Load}
    \vspace{-0.9\baselineskip}
    
\end{figure*}
}%

In \cref{fig:T0W0_ModelError}, both CMs show increasing $|d_z|$ and a more negative $\Delta$Pitch from P1 to P4, consistent with growing \emph{in-plane} deformation relative to the CC model. For an upside-down, tendon-driven CM, self-weight acts as a persistent in-plane load, introducing a systematic bias in $(d_x,d_z,\Delta\text{Pitch})$. The near-overlap of NiTi results between T0 and T100 suggests DW has a weaker effect on in-plane deformation than on the out-of-plane channels $(\Delta\text{Yaw},d_y)$ and $\Delta$Roll. In contrast, the U-joint CM shows larger in-plane model mismatch as expected(more negative $d_z$ and $\Delta$Pitch), indicating stronger gravity-driven deviation from the CC assumption.

Overall, under peak DW the U-joint CM achieves approximately $2.5$--$4\times$ lower yaw error, $2.5$--$45\times$ lower $d_y$, and up to $\sim 5\times$ lower roll error than the NiTi CM. The consistently lower and similar $\Delta$Yaw, $d_y$ and $\Delta$Roll with lower SD across T0 and T100  confirms stronger resistance to out-of-plane bending for the U-joint design. Although the U-joint CM shows larger in-plane error, its SD remains bounded and the T0 and T100 trends align closely, including at the extreme point P4, compared with the NiTi CM.



\subsection{Case (ii): Payload-induced pose variation of the U-joint CM in still air (baseline: Case~(i), $T=0$)}
\vspace{-7pt}
\label{sec: T0_W0-W50}

Here, Results are reported relative to the U-joint T0\_W0 baseline. The out-of-plane signatures remain tightly bounded under EE loading. Across W10--W50, $\Delta$Yaw stays within $\sim\pm 4^\circ$ and $d_y$ within $\sim\pm 4$\,mm over the workspace ($\approx 1\%$ of the CM length), indicating strong resistance to payload-induced out-of-plane bending. Twisting is also limited: $\Delta$Roll remains bounded with a small spread across payloads. The largest variability occurs at W50, with SD of $\sim\pm 2^\circ$ in yaw and $\sim\pm 1.5$\,mm in $d_y$ ($\approx 0.4\%$ of the CM length). A single outlier at W20 appears in $\Delta$Roll, likely due to an experimental disturbance. Overall, these results confirm that the U-joint profile maintains out-of-plane robustness under load.

AS expected, while payload increases (W10--W50), the in-plane signatures $(d_x,d_z,\Delta\mathrm{Pitch})$ grow: $d_z$ and $\Delta$Pitch become more negative and peak at P4, indicating gravity-driven sag ($d_z$ shift $\approx 18\%$ of CM length). $d_x$ also shifts negative, peaking at P3 ($\approx 13\%$), capturing the coupled in-plane offset. Despite these biases, repeatability remains consistent with SD: $<10$\,mm and $<2^\circ$. These systematic trends can be exploited to tune U-joint CAAMS kinematics via MPC or data-driven control using the experiment data obtained.


{%
\begin{figure}[htb]
    \vspace{6pt}
    \centering
    \includegraphics[scale=1]{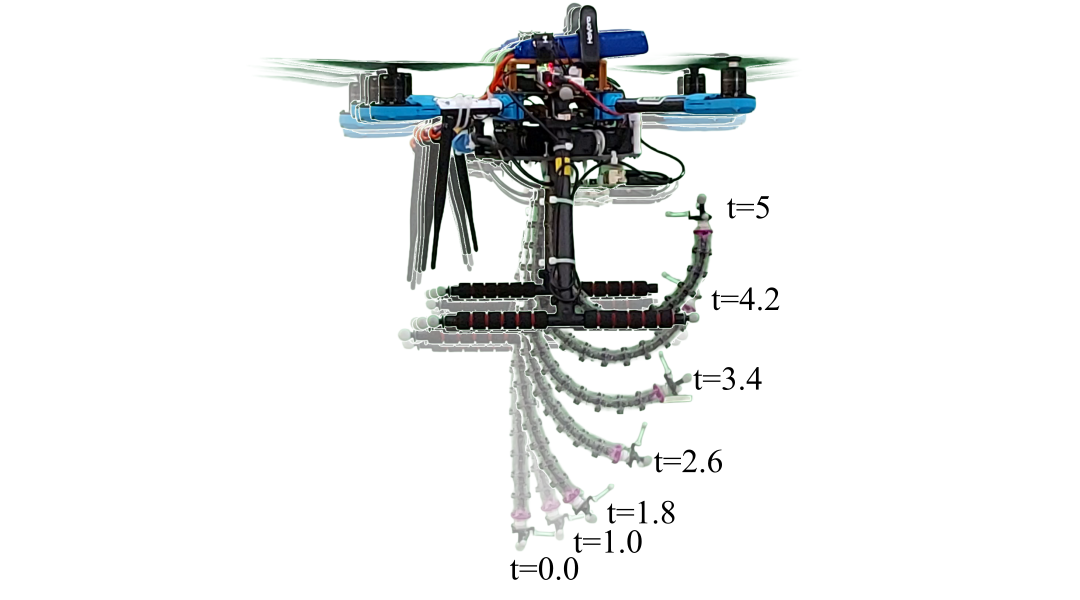}
    \caption{Time-lapse composite of the U-joint CAAMS during the hover test at $T=20$\,s and $W=0$, showing CM motion over a quarter-cycle. }
    \label{fig:timelapse}
    \vspace{-6pt}
\end{figure}
}%

\subsection{Case (iii): Payload and DW-induced pose variation of the U-joint CM in under extreme DW (baseline: Case~(i), $T=100\%$)}

Fig. 7 shows T100 load sweeps relative to the T0\_W0 baseline. The out-of-plane signatures remain bounded: $\Delta$Yaw stays within a few degrees (within the $\pm 2.5^\circ$ axis range) and $d_y$ within $\sim 4$ to $-8$\,mm ($\approx 3\%$ of CM length), with slightly larger magnitudes at P3--P4. Unlike Case~(ii), $\Delta$Yaw and $d_y$ take mixed signs, consistent with DW-induced loading. $\Delta$Roll is also limited (typically $\sim 0$ to $-5^\circ$), peaking at P4 for higher payloads. Overall, the bounded $(\Delta\text{Yaw},d_y)$ and $\Delta$Roll confirm robust out-of-plane performance under DW and payload.

Under DW, the in-plane channels $(d_x,d_z,\Delta\mathrm{Pitch})$ remain load and workspace-dependent. With increasing payload, $d_z$ and $\Delta$Pitch become more negative and peak at P4, indicating gravity-driven sag and the DW load.  $d_x$ also shifts negative, peaking near P3 and partially recovering toward P4, consistent with a coupled in-plane offset. 
However the pattern remains similar between case(ii) and case(iii). Further tight error bounds and small SD confirm that the U-Joint CM improves the stability as required by the established design intention under the Section II.


\subsection{Flight hover stability test}
We follow the CPD hover protocol in \cite{xu2025atom}, but do not perform a direct  comparison due to differences in platform and manipulator morphology. \cref{fig:timelapse} shows a time-lapse composite of the U-joint CAAMS over a quarter of the slowest actuation cycle ($T=20$\,s).
\Cref{fig:hoverTest} plot reports hover-position RMSE during U-Joint CM actuation for W0, W30, and W50 over cycle times $T\in\{2,4,8,12,16,20\}$\,s; hover without CM actuation defines the baseline, with RMSE/Mean/Max overlaid on the plot.

{%
\begin{figure}[htb]
    \vspace{6pt}
    \centering
    \includegraphics[width = \linewidth]{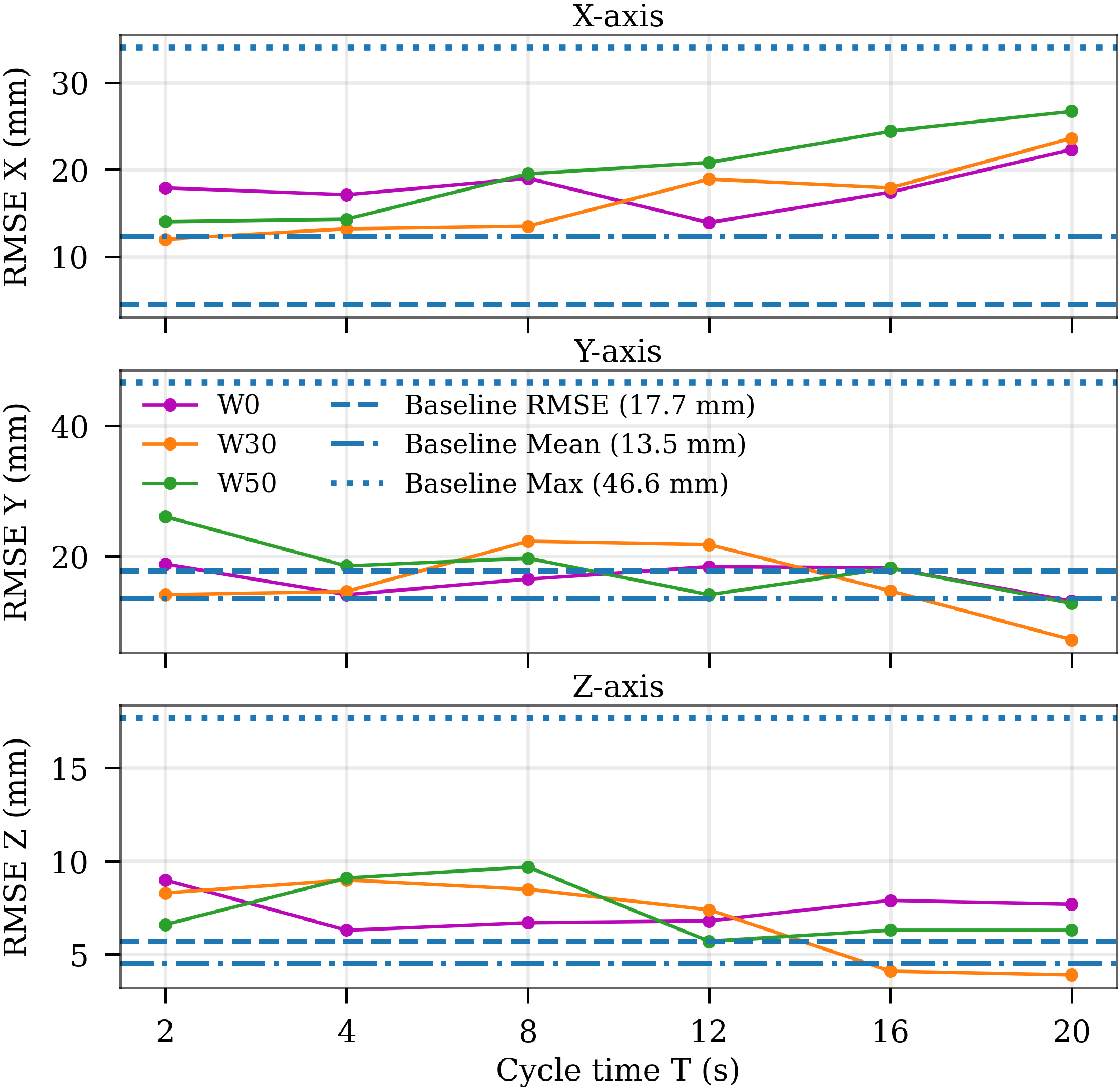}
    \caption{RMSE of Cartesian positions during CPD hover tests with sinusoidal U-joint CM actuation. Results are shown for payloads W0, W30, and W50 over six cycle times ($T\in\{2,4,8,12,16,20\}$\,s;  Baseline hover RMSE/Mean/Max (no actuation) are overlaid for reference. }
    \label{fig:hoverTest}
    \vspace{-6pt}
\end{figure}
}%

CPD shows axis-dependency: the $x$-axis (CM actuation direction) is most sensitive consistent with \cite{xu2025atom}, and deviates most from the baseline mean at slow actuation ($T=20$\,s). The Max $x$-axis errors are 52.9\,mm (at 50\,g), 47.9\,mm (at 30\,g), and 47.9\,mm (at 20\,g). In all other cases, RMSE remains close to the baseline mean. The $y$- and $x$-axis RMSE are  bounded within 10--25\,mm and 3--10\,mm, respectively, and the $z$-axis error is the smallest while being most sensitive to fast actuation. Overall, RMSE remains bounded across all payloads for $T\in\{2,4,8,12\}$\,s on all axes, indicating stable hover and bounded range CPD during U-joint CM actuation at corresponding speeds. These results support the applicability of the U-joint CAAMS for real-world deployment. In this case, We analyzed one representative bending plane; a full 360$^\circ$ bending plane sweep is left for future real-world validation.

\subsection{Application experimentation of U-Joint CAAMS}
Guided by the CPD hover tests, we conducted three application flights with the U-joint CAAMS (Fig.~1(b)--(d)): (1) water sampling, (2) spot spraying on an artificial plant target, and (3) whole-body-grasp object transport of a 780\,g tripod (see the accompanying video\textsuperscript{$\dagger$}). For the transport task, the pump was removed to reduce system mass. The U-joint CAAMS was remotely operated under position control mode launched from the GCS and successfully performed water sampling, spot spraying and object transportation tasks. 

Overall, the laboratory demonstrations support the purpose of the new design. 
The water sampling and spraying demonstrations used a low-flow peristaltic pump (85\,ml/min) for the proof-of-concept; practical deployments may require higher-flow and higher-pressure pumps. We also observed strong DW disturbance of the water sampling surface and sprayed target, which can degrade sampling/spraying quality. To reduce disturbance at the interaction site, the U-joint CM can be integrated with Tilt-X's tilting and telescopic extension to operate outside the DW region \cite{uthayasooriyan2026tiltxenablingcompliantaerial}. During whole-body grasping, the U-joint CM deformed and deviated from its initial shape. For object transport, the observed shape deformation is acceptable and reflects the its improved payload capacity. Finally, the CPD results and successful demonstrations indicate that closed-loop control and autonomous operation are feasible next steps for the U-joint CAAMS.

\section{conclusions}


This paper presented a tubular universal-joint CM profile for aerial manipulation with an integrated conduit for protected routing and fluid delivery. Bench tests in still air and propeller DW showed improved resistance to payload- and DW-induced out-of-plane bending compared with the Tilt-X continuum design. The repeatable load and DW-dependent biases suggested that a learned correction model could be integrated within an MPC framework. We quantified hover stability via in-flight CPD analysis over payload and actuation-speed sweeps, and demonstrated the applicability of the system through in-flight water sampling, spot spraying, and object transport. To improve kinematic accuracy, future work will consider a two-section design and higher-fidelity modeling. The current system operates open loop; IMU-based feedback control, as in \cite{peng2025dexterous,peng2023aecom}, can compensate unmodeled effects beyond CC modeling and mitigate mechanical disturbances. Further integration of improved U-Joint CM design with Tilt-X is expected to improve accuracy of environmental interaction in the presence of downwash.

\bibliographystyle{IEEEtran} 
\bibliography{references}

\end{document}